\documentclass{article}
\usepackage{spconf,amsmath,graphicx,algorithm}
\usepackage[noend]{algpseudocode}
\usepackage{multirow}

\def\SessionLog{{EventLog}}
\def\Lag{{TL}}
\def\LagExpanded{{Translation Lag}}
\newcommand\bias[0]{\beta}

\newcommand*\samethanks[1][\value{footnote}]{\footnotemark[#1]}

\title{RE-TRANSLATION STRATEGIES FOR LONG FORM, SIMULTANEOUS, SPOKEN LANGUAGE TRANSLATION}
\name{Naveen Arivazhagan\sthanks{Equal contributions}, Colin Cherry\samethanks, Te I, Wolfgang Macherey, Pallavi Baljekar and George Foster}
\address{Google Research}
\begin{document}
%
\maketitle
\begin{abstract}
We investigate the problem of simultaneous machine translation of long-form speech content. We target a continuous speech-to-text scenario, generating translated captions for a live audio feed, such as a lecture or play-by-play commentary. As this scenario allows for revisions to our incremental translations, we adopt a re-translation approach to simultaneous translation, where the source is repeatedly translated from scratch as it grows. This approach naturally exhibits very low latency and high final quality, but at the cost of incremental instability as the output is continuously refined. We experiment with a pipeline of industry-grade speech recognition and translation tools, augmented with simple inference heuristics to improve stability. We use TED Talks as a source of multilingual test data, developing our techniques on  English-to-German spoken language translation. Our minimalist approach to simultaneous translation allows us to scale our final evaluation to several other target languages, dramatically improving incremental stability for all of them.
\end{abstract}
\section{Introduction}
\label{sec:intro}
Recent advances in neural speech and text processing have greatly improved speech translation and even simultaneous speech translation. Many such improvements take the form of specialized models and training procedures which operate under a \emph{streaming} constraint, where each update to the translation must extend the previous translation. However, when the output modality for the translation is text (rather than, e.g., speech), we can make occasional revisions to previous predictions without being too disruptive to the user. 

The ability to revise previous partial translations makes simply re-translating each successive source prefix a viable strategy. This approach has the advantage of low latency, since it always attempts a translation of the complete source prefix, and high final-translation quality, since it is free to generate its final translation with knowledge of the full source sentence. The freedom to make revisions poses challenges for evaluation, though. We need to control key aspects of system performance: quality, latency, and stability. Unstable translations make many revisions as more source content arrives, and the display \emph{flickers} with each change. If the revision rate is very high, having low latency is pointless; on the other hand, if latency is very high, performing simultaneous translation is pointless. We propose an evaluation framework that measures each of these factors separately, and use it to characterize the approaches we consider.

Equipped with this evaluation framework, we develop a re-translation approach augmented with some simple yet effective inference modifications to neural machine translation (NMT). We demonstrate that even with off-the-shelf ASR and NMT modules that are in no way fine-tuned for speech-translation, we are able to provide high quality speech-to-text translation on arbitrarily long unsegmented audio. By virtue of being built on top of generic ASR and NMT models, our approach is easy to maintain and immediately benefits from improvements to its underlying components with no extra overhead. We demonstrate
the robustness
of this approach by reporting simultaneous speech-translation results on several target languages,
only one of which we used for development.





\section{Background}
\label{sec:background}
Initial work on simultaneous translation was centered around dividing transcribed text into translation units that are translated independently~\cite{Fugen2007,Bangalore2012}. Translation results are then concatenated and cannot be subsequently revised. Our re-translation approach is similar only in that it does not require specialized NMT models.

The line of work most relevant to ours is \cite{niehues2016dynamic,Niehues2018}, which contains the only systematic studies of re-translation for simultaneous translation that we know of. We build directly upon the re-translation strategy proposed in \cite{Niehues2018}. We extend their work along several fronts, starting by obtaining a more holistic view of  system performance by tracking latency, which allows for comparisons between systems that translate at different speeds. With a novel inference (search) algorithm carried out during re-translation, we are able to reduce the instability inherent in such re-translation approaches by up to 100x.

Much recent work on simultaneous translation has focused on developing streaming systems that never revise~\cite{Grissom2014,Gu2017,Press2018,Ma2019,Avari2019,Xiong2019}. This is typically achieved by developing specialized models to translate prefixes of sentences with an implicit or explicit agent that decides when to stop translating and instead wait for more source tokens. With the exception of \cite{Xiong2019}, all of these were evaluated on written source text, while we evaluate on spoken language, and additionally permit output to be revised.

\section{Evaluation Framework}
\label{sec:evaluation_framework}
One of our main contributions is the establishment of an evaluation framework suitable for long-form spoken language re-translation.
This necessitates measuring quality, latency, and stability.
The guiding principle is to operate at the document level wherever possible.

At the core of our framework is the \emph{\SessionLog{}}, an ordered list of {\em events}, where the $i^{\mathit{th}}$ event is a triple consisting of:
\begin{itemize}
    \item $s_i$ the source text recognized so far,
    \item $o_i$ the translated output text currently displayed, and
    \item $t_i$ a time-stamp.
\end{itemize}
Events are logged every time the source or output changes. 
Crucially, each event $i$ records the entire session, or spoken document, up to time $t_i$, allowing us to work and evaluate without providing our system gold-standard segmentations.
Note that each event can arbitrarily alter the previously displayed text.
Text is tokenized into token vectors $\mathbf{s}_i=[\mathbf{s}_{i,1} \ldots \mathbf{s}_{i,j} \ldots \mathbf{s}_{i,|s_i|}]$ and $\mathbf{o}_i=[\mathbf{o}_{i,1} \ldots \mathbf{o}_{i,j} \ldots \mathbf{o}_{i,|o_i|}]$ where the $|\cdot|$ operator measures the length of a vector.
We use $I$ to represent the total number of events.
An example \SessionLog{} is shown in Table~\ref{tab:session}.
\begin{table*}[t]
\begin{center}
\begin{tabular}{|c|l|l|}
\hline
Time & Source & Output \\
\hline
2.0 & Neue Arzneimittel k\"onnten & New Medicines  \\
3.5 & Neue Arzneimittel k\"onnten Eierstockkrebs & New Medicines may be ovarian cancer \\
4.2 & Neue Arzneimittel k\"onnten Eierstockkrebs verlangsamen & New Medicines may slow ovarian cancer \\
\hline
\end{tabular}
\end{center}
\caption{An example \SessionLog{} for the tokenized German sentence, 
``Neue Arzneimittel k\"onnten Eierstockkrebs verlangsamen'',
with reference, ``New drugs may slow ovarian cancer''.}
\label{tab:session}
\end{table*}

\subsection{Quality}
\label{sec:quality}
We only measure the quality of the translation of the final output vector $\mathbf{o}_I$. We measure case-sensitive BLEU~\cite{Papineni2002} after aligning the unsegmented translation output with the provided reference sentences by minimizing WER~\cite{Matusov2005}.

Note that we do not explicitly measure the translation quality of intermediate events. Instead, this is implicitly captured by our metrics for latency and stability, which indicate whether intermediate events differed significantly from the final output.

\subsection{Latency}
We propose to measure latency relative to the speaker based on the time delay between when a source word was spoken versus when a corresponding output word was finalized. A word is finalized in the first event where the word and any words before it remain unchanged for all subsequent events. Formally, the finalization event index for the $j^{\mathit{th}}$ word ($1 \leq j \leq |\mathbf{o}_I|$) in the final output vector $\mathbf{o}_I$ is given by:
\begin{equation}
f(j)= \min_i \textrm{ s.t. } \mathbf{o}_{i', j'} = \mathbf{o}_{I, j'} \; \forall i' \geq i \; \mathrm{and} \;  \forall j' \leq j
\end{equation}
and its finalization time is $t_{f(j)}$.

\def\sysSent{{\mathbf{u}}}
\def\refSent{{\mathbf{v}}}
We impose a correspondence between source and output words based on their relative positions in a heuristically-derived parallel sentence pair. We use the same WER alignment as in $\S$\ref{sec:quality} to align $\mathbf{o}_I$ to the segments of the reference translation, which in turn have a provided segment alignment to the reference source transcription. Let $\sysSent$ be the vector of absolute positions in the output segment containing $j$, and $\refSent$ be the vector of positions in its parallel reference source segment. We define the reference source position $j^*$ for the $j^{\mathit{th}}$ output token as $j^*= (j - \sysSent_0) \cdot|\refSent|/|\sysSent| + \refSent_0$.\footnote{We originally tried correspondence based on relative document positions: $j^*=j \cdot |\mathbf{s}_I| / |\mathbf{o}_I|$, but we found that this $j^*$ could appear tens of tokens before or after the actual source token being translated at position $j$.} 
While using a word-aligner~\cite{shavarani2015learning} would capture semantic correspondence, we opt instead for this simpler and parameter-free temporal correspondence.
The latency for the $j^{\mathit{th}}$ output token is then:
\begin{equation}
\mathrm{\Lag{}}(j) = t_{f({j})} - \mathit{time(j^*)}
\end{equation}
where $\mathit{time}(\cdot)$ gives the utterance time for a reference source token.
We average across all output tokens to get the \LagExpanded{} (\Lag{}) for the \SessionLog{}:
$\mathrm{\Lag{}} =  \frac{1}{|\mathbf{o}_I|}\sum_{j=1}^{|\mathbf{o}_I|} \mathrm{\Lag{}}(j)$.
This measures in seconds how target content lags behind spoken source content.

\subsection{Stability}
A live translation system that is allowed to make revisions could exploit this to frequently make risky guesses that would lower lag on the off-chance that they were correct. 
These frequent revisions would produce a visible flickering effect that can be irritating and distracting to users.
Following \cite{niehues2016dynamic, Niehues2018}, we measure flicker
directly with what we dub \textbf{erasure} (E), which
measures the number of tokens that must be deleted from the suffix of the previous
translation to produce the next.
For the $i^{\mathit{th}}$ event, the erasure of the output is defined as:
\begin{equation}
\mathrm{E}(i) = |\mathbf{o}_{i-1}| - |\mathrm{LCP}(\mathbf{o}_i, \mathbf{o}_{i-1})|
\end{equation}
where $\mathrm{LCP}$ calculates the longest common prefix of two sequences.
For example, in Table~\ref{tab:session}, output at time 4.2 replaces \textit{be} with 
\textit{slow}, resulting in an erasure of 3 for the deletion of the suffix 
\textit{be ovarian cancer}.
We now define the normalized-erasure of an \SessionLog{} as the aggregate erasure divided by the output length in the final event: $\mathrm{NE} = \frac{1}{|\mathbf{o}_I|}\sum_i^I E(i)$.

\section{System Description}
\label{sec:sysdesc}
The guiding principle for our current effort in speech translation is to redevelop as few components as possible. Instead we prefer to fallback to generic, readily available, and highly reliable ASR and NMT systems. Our ASR module performs live transcription and punctuation. The NMT system is trained on large amounts of parallel sentences mined from the web and is considered to be a general domain translation model. It is not fine-tuned  for the speech domain or simultaneous translation. Our goal here is to explore the maximum potential of such off-the-shelf systems for live speech translation.

\subsection{Combining Independent ASR and NMT}
To enable long-form speech translation, our system maintains state as time progresses. The state consists of the transcription and translation of the audio stream received so far. As new bytes stream in, they are sent to the ASR system which emits transcribed text in real-time as and when it becomes sufficiently confident. The emitted transcription results are appended to the continuously growing transcription document. On every such update, we take the last, potentially incomplete sentence from the transcription document and send it to the NMT model for translation. The result from the NMT system is appended to the translation document, replacing a previous translation of the same sentence if it was present.


\subsection{Biasing Beam Search for NMT}
To reduce flicker, we may bias a re-translation model to respect its previous predictions. Occasionally the changes that a model makes are necessary fixes that are needed to improve the quality of the translation. Other times, they are just superficial changes that do not improve quality or understanding. With biased beam search, we trade quality to improve stability, as we bias the system toward its earlier decisions that were made with less source context, hoping to eliminate only superficial changes.

When decoding a source prefix, we not only pass in the tokens of the source, but additionally provide the decoded target of the previous source prefix. We modify beam search to interpolate between the posterior distribution provided by the NMT model and the degenerate one hot distribution formed by the provided target. Let $y'$ be the translation output for the previous hypothesis, then for the current hypothesis, the probability of the next token is:
\[
p'(y_j|y_{<j},x_{\le i}) = (1-\bias)\cdot p(y_j|y_{<j},x_{\le i}) + \bias\cdot\delta(y_j, y'_j)
\]
We only bias a beam for as long it has strictly followed the target. After the first point of divergence, no bias is applied, and only the posterior from the NMT model is used.

\subsection{Masking Unstable Tokens}
Much of the flicker in re-translation happens toward the end of the system output. This is because these latest target tokens are more likely to have dependencies on  source tokens that have yet to arrive.
If a system simply waits for the arrival of said source tokens, it can reduce
flicker at the cost of increased latency.

Since they do not have the option to flicker, all streaming systems have some form of agent that decides how long they should wait for more source words. \cite{Gu2017} and \cite{Avari2019} learn this agent, while \cite{Cho16} implement the agent as heuristics over the NMT model state. Recently, \cite{Ma2019} showed that a very simple wait-$k$ heuristic can also be very effective. 
Inspired by wait-$k$, we adopt a simple heuristic at inference time where we mask the last $k$ tokens of the predicted target sentence. The masking is only applied if the current source (and therefore the target) are prefixes and not yet completed sentences. We refer to this heuristic as mask-$k$.

\subsection{Evaluation}
We evaluate our system on TED talk data, selecting talks based on test sets from past IWSLT spoken language translation evaluations. For dev, we use the talks in tst2015~\cite{Cettolo2015_iwslt} and for test we use the talks in tst2018~\cite{Niehues2018_iwslt}. We do not use the transcripts or any sentence segmentation provided in the official IWSLT shared task, as we found that these frequently omitted transcriptions for several sentences. Instead, we scrape raw transcripts and translations into different languages directly from ted.com,\footnote{\texttt{ted.com/talks/subtitles/id/\$TALK\_ID/lang/\$LANG}} without further post-processing. We derive token-level utterance times by linearly interpolating the TED's closed-captioning timings. Using raw TED data has the advantage of allowing us to test any target language for which translations are available on ted.com. For each TED talk, we stream the full roughly 10 minute long audio file to our system without doing any segmentation. 

\section{Experiments}
\label{sec:experiments}
\begin{table}[t]
    \centering
    \begin{tabular}{lcc|ccc}
        System      & $\bias$ & $k$ & BLEU  & \Lag{} & NE  \\ \hline
        Baseline  & 0.0 & 0 & 20.40 & 4.13 & 2.11  \\
         + Bias   & 0.5 & 0 & 20.03 & 3.00 & 0.72 \\
         + Mask-k & 0.0 & 10 & 20.40 & 5.98 & 0.53 \\
         + Both   & 0.5 & 5& 20.17 & 4.11 & 0.12 \\
    \end{tabular}
    \caption{English-to-German results on our TED test set. \Lag{} is \LagExpanded{}, measured in seconds. NE is Normalized Erasure, measured in number of erased partial target tokens per final target token.}
    \label{tab:EnDeResult}
\end{table}

We first study the effectiveness of our proposed NMT heuristics in our English-to-German TED Talk simultaneous spoken language translation scenario. Test set results, measured in BLEU, \LagExpanded{} (\Lag{}) and Normalized Erasure (NE), are shown in Table~\ref{tab:EnDeResult}. Our baseline is a basic re-translation system with standard beam search. Its \Lag{} indicates that its stable predictions lag 4.13 seconds behind the spoken source, while its NE shows that 2.11 tokens are incrementally erased and replaced for every token in the final translation -- the translation is constantly in flux.

We now add biased beam search alone (\emph{+Bias}), then masking alone (\emph{+Mask-k}), and then both.
The hyper-parameters, $\bias \in \{0.1, 0.2, 0.3, 0.5 0.8, 1.0\}$ and $k \in \{0, 1, 2, 3, 4, 5, 7, 10\}$, are tuned on the development set for each configuration. In the presence of multiple objectives, we follow a straightforward strategy wherein we optimize one metric while placing constraints on the others. For \emph{+Bias}, we vary $\bias$ to minimize NE while staying within 1 BLEU of Baseline. Since words are finalized sooner with bias, \Lag{} also decreases. For \emph{+Mask-k} we show results for a configuration that lags the baseline by 2 seconds but lowers NE. As expected, BLEU is not affected by masking alone. Finally, in \emph{+Both}, we vary both $\bias$ and $k$ to maximize BLEU, while keeping \Lag{} less than that of the baseline, and NE less than the best NE for either heuristic alone.
Using both strategies allows us to reach an excellent compromise: comparable BLEU and \Lag{} to the baseline, while producing a 20x reduction in NE.

\begin{figure}[t]
    \centering
    \includegraphics[width=0.45\textwidth]{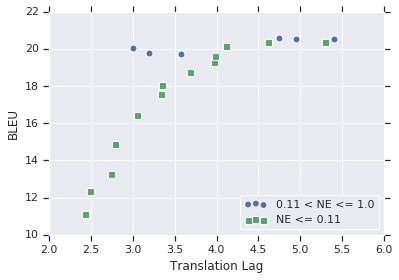}
    \caption{BLEU-versus-Lag curves on our English-to-German TED data, corresponding to dev set Pareto frontiers for two Erasure ranges, projected onto the test set.}
    \label{fig:EnDeCurve}
\end{figure}

To visualize a larger range of quality-latency-stability trade-offs, we construct Pareto-optimal quality-versus-latency curves for two NE constraints in Figure~\ref{fig:EnDeCurve}.
The constraints correspond to moderate erasure (NE $\leq$ 1.0) and low erasure (NE $\leq$ 0.11 -- chosen to include our \emph{+Both system}).
The systems to be included in each curve are determined on the development set, but we graph their test set scores.
As one can see, at low levels of erasure, quality falls off quickly as \Lag{} drops below 4.
If we want to achieve a low \Lag{} without sacrificing too much quality, we have to incur a moderate amount of flicker.

We use TED talks tst2018 as a source of multilingual test data to demonstrate the scalability of our re-translation model, and show the results in Table~\ref{tab:ScalabilityResult}. 
Without any further fine-tuning, we use the same hyper-parameters from our English-to-German experiments ($\bias=0.5$ and $k=5$) to reduce erasure and lag. 
Our approach is remarkably robust, never reducing BLEU by more than 1 point, consistently improving lag, and always reducing erasure to negligible amounts.
\begin{table}[t]
    \centering
    \begin{tabular}{ccc|ccc}
        Language & $\bias$ & $k$ & BLEU  & \Lag{}  & NE   \\ \hline \hline
        \multirow{2}{*}{De}       & 0.0  & 0 & 20.40 & 4.13 & 2.11 \\
                 & 0.5  & 5 & 20.17 & 4.11 & 0.12 \\ \hline
        \multirow{2}{*}{Es}       & 0.0  & 0 & 24.94 & 6.00 & 4.34 \\
                 & 0.5  & 5 & 25.32 & 4.24 & 0.05 \\ \hline
        \multirow{2}{*}{Fr}       & 0.0  & 0 & 21.36 & 5.96 & 5.25 \\
                 & 0.5  & 5 & 21.02 & 3.86 & 0.11 \\ \hline
        \multirow{2}{*}{It}       & 0.0  & 0 & 25.34 & 4.51 & 2.08 \\
                 & 0.5  & 5 & 25.19 & 3.95 & 0.01 \\ \hline
        \multirow{2}{*}{Pt}       & 0.0  & 0 & 25.11 & 4.77 & 2.31 \\
                 & 0.5  & 5 & 24.39 & 4.02 & 0.01 \\ \hline
        \multirow{2}{*}{Nl}       & 0.0  & 0 & 22.00 & 4.41 & 2.14 \\
                 & 0.5  & 5 & 21.85 & 4.05 & 0.01 \\ \hline
        \multirow{2}{*}{Ru}       & 0.0  & 0 & 17.35 & 4.95 & 2.76 \\
                 & 0.5  & 5 & 16.97 & 4.11 & 0.01 
    \end{tabular}
    \caption{We demonstrate the scalability of our approach by evaluating translation to many languages on our TED test set. Hyper-parameters, $\bias$ and $k$, were only tuned on the TED German dev set.}
    \label{tab:ScalabilityResult}
\end{table}

\section{Conclusion}
\label{sec:conclusion}
We develop an evaluation framework that enables us to comprehensively measure the performance of simultaneous translation systems. Under this framework, we propose and demonstrate simple inference algorithms that can be used to stitch together off-the-shelf ASR and NMT models to obtain high quality simultaneous translation. By eschewing the development of specialized models from scratch, we are able to provide support for simultaneous translation to any language for which we have a translation system. We believe the simplicity of our approach will enable it to serve as an important baseline for subsequent work on simultaneous translation.

\bibliographystyle{IEEEbib}
\bibliography{refs}

\end{document}